\documentclass[letterpaper, 10 pt, conference]{ieeeconf}  

\IEEEoverridecommandlockouts                              

\usepackage{xcolor}
\usepackage{amsmath} 
\usepackage{amssymb}  
\usepackage{array}
\usepackage{booktabs}
\usepackage{graphicx,nicefrac}
\graphicspath{ {images/} }
\usepackage[normalsize,font=footnotesize]{caption}
\usepackage{xspace}
\usepackage{algorithm}
\usepackage{hyperref}
\usepackage{balance}

\let\labelindent\relax
\usepackage{enumitem}

\usepackage[style=ieee,backend=bibtex,isbn=false,doi=false,url=false,dashed=false]{biblatex} 
\usepackage{microtype}
\renewcommand{\baselinestretch}{0.94}

\usepackage{multirow}
\usepackage{colortbl}

\newcommand{\theModel}{CF\nobreakdash-EVRP\xspace}
\newcommand{\theAlgorithm}{ComSat\xspace}
\newcommand{\MonoMod}{MonoMod\xspace}

\newcommand{\OR}{\ensuremath{\mathit{OR}}\xspace}
\newcommand{\C}{\ensuremath{\mathit{C}}\xspace}
\newcommand{\D}{\ensuremath{\mathit{D}}\xspace}
\newcommand{\Origin}{\ensuremath{\mathit{O}}\xspace}

\newcommand{\jobSet}{\textrm{$ \mathcal{J}$}}
\newcommand{\taskSet}{\textrm{$\mathcal{K}$}}
\newcommand{\precTask}{\textrm{$\mathcal{P}$}}
\newcommand{\nodeSet}{\textrm{$\mathcal{N}$}}
\newcommand{\vehicleSet}{\textrm{$\mathcal{V}$}}
\newcommand{\edgeSet}{\textrm{$\mathcal{E}$}}
\newcommand{\Path}{\textrm{\emph{path}}}
\newcommand{\Perm}{\textrm{\emph{Perm}}}
\newcommand{\Start}{\textrm{\emph{start}}}
\newcommand{\Routes}{\textrm{\emph{Routes}}}
\newcommand{\late}{\textrm{\emph{late}}}
\newcommand{\length}{\textrm{\emph{length}}}
\newcommand{\allo}{\textrm{\emph{allo}}}
\newcommand{\End}{\textrm{\emph{end}}}
\newcommand{\len}{\textrm{\emph{len}}}
\newcommand{\node}{\textrm{\emph{node}}}
\newcommand{\edge}{\textrm{\emph{edge}}}

\title{\LARGE \bf
An SMT Based Compositional Algorithm to Solve a Conflict-Free Electric Vehicle Routing Problem
}

\author{Sabino Francesco Roselli$^{1}$ and Martin Fabian$^{1}$ and Knut \AA kesson$^{1}$     
\thanks{We gratefully acknowledge financial support from Chalmers AI Research Centre (CHAIR), AB Volvo (Project ViMCoR), the support from Per-Lage Götvall at Volvo Group Truck Operation, and the Wallenberg AI, Autonomous Systems and Software program (WASP) funded by the Knut and Alice Wallenberg Foundation.
$^{1}$Department Electrical Engineering, Chalmers University of Technology,
        G\"oteborg, Sweden
        {\tt\small \{rsabino, fabian, knut\}@chalmers.se}}%
}

\begin{document}

\maketitle
\thispagestyle{empty}
\pagestyle{empty} 

\begin{abstract}

The Vehicle Routing Problem (VRP) is the combinatorial optimization problem of designing routes for vehicles to visit customers in such a fashion that a cost function, typically the number of vehicles, or the total travelled distance is minimized. The problem finds applications in industrial scenarios, for example where Automated Guided Vehicles run through the plant to deliver components from the warehouse. This specific problem, henceforth called the Electric Conflict-Free Vehicle Routing Problem (\theModel), involves  constraints such as limited operating range of the vehicles, time windows on the delivery to the customers, and limited capacity on the number of vehicles the road segments can accommodate at the same time. Such a complex system results in a large model that cannot easily be solved to optimality in reasonable  time. We therefore developed a compositional algorithm that breaks down the problem into smaller and simpler sub-problems and provides sub-optimal, feasible solutions to the original problem. The algorithm exploits the strengths of SMT solvers, which proved in our previous work to be an efficient approach to deal with scheduling problems. 
Compared to a monolithic model for the \theModel, written in the SMT standard language and solved using a state-of-the-art SMT solver the compositional algorithm was found to be significantly faster.

\end{abstract}

\section{Introduction} \label{sec:intro}

The use of Automated Guided Vehicles (AGVs) for \emph{just-in-time} deliveries is becoming  common  in modern manufacturing facilities \cite{azadeh2017robotized}. Adopting this solution, rather than storing all the components by the assembly line, makes the environment more worker-friendly and using AGVs instead of fixed transportation belts (or similar) makes it more flexible \cite{theunissen2018smart}. 

A system like this  requires a Scheduler that guarantees that
the deadlines for the delivery of components are met, but also that AGVs do not create queues by the workstations by arriving too early. Also, AGVs are battery-powered so their operating range is limited. Therefore they need to recharge 
occasionally,
and this must be taken into account when designing the schedule. Finally, though AGVs are usually equipped with low-level controllers to avoid dangerous conditions, they may not be able to avoid deadlocks, i.e. getting stuck
due to circular waiting between them.
Hence the scheduler must ensure such situations are avoided. 

This type of problem can be modelled as a Vehicle Routing Problem (VRP) \cite{braekers2016vehicle}, the combinatorial optimization problem of designing routes for vehicles to visit customers, such that a cost function is optimized. There exist extensions of the VRP that involve additional constraints, such as time windows for the customers' service (VRP with time windows, or VRPTW~\cite{desrochers1992new}), limited operating range of the vehicles and possibility to recharge at the charging stations (Electric VRP, or E-VRP~\cite{schneider2014electric}), and limitations on the capacity of the road segments that vehicles drive on (dispatch and conflict-free routing problem (DCFRP)~\cite{krishnamurthy1993developing}). The problem we are tackling in this work, henceforth called the Electric Conflict-Free Vehicle Routing Problem (\theModel), involves all these features and also additional constraints related to the customers' service. 

For relatively small size problem instances, mixed integer linear programming solvers (MILP, \cite{brahimi2016multi}) can provide good solutions to VRPs in a short time. However, for larger problems, MILP solvers are often not fast enough and specific-purpose algorithms involving local search~\cite{braysy2005vehicle}, column generation~\cite{riazi2019scheduling} or stochastic methods~\cite{baker2003genetic, gong2011optimizing} are used instead. Recent work focusing on fleets of electric vehicles~\cite{rossi2019interaction}, as well as conflict-free routing~\cite{thanos2019dispatch} show applications of such approaches to real-world problems.

In our previous work \cite{roselli2021smt}, we presented a monolithic formulation (\MonoMod) to model the \theModel; based on our previous findings from \cite{roselli2018smt}, we decided to formulate the model in SMT standard Language (Satisfiability Modulo Theory, \cite{barret_2009,DeMoura_2011}) and solve it using the state-of-the-art SMT solver Z3 \cite{de_moura_2010}; as expected, our approach was not able to solve large problem instances in reasonable time. 
In fact, though Z3 has proven very efficient in solving combinatorial optimization problems \cite{bjorner_2015}, our formulation quickly leads to a state-space explosion as the number of vehicles and jobs, and the time horizon (a fixed point of time in the future when certain processes will be evaluated or assumed to end) for the jobs' execution increases. This is mainly due to the necessity of discretizing time in order to keep track of the vehicle's locations and avoid collisions. The compositional algorithm (\theAlgorithm) we present in this work breaks down the \theModel into sub-problems so that time discretization can be avoided and a feasible solution can be reached quickly. The algorithm exploits the strengths of the SMT solvers by reducing the problem to smaller Job Shop Problems~\cite{carlier1989algorithm} (JSPs). It first selects a set of paths to uniquely connect any two customers, since in a real plant there can exists multiple paths; it then solves a VRP to design routes to serve all customers within their time windows; if such a solution exists, it matches some of the available vehicles with the generated routes; finally, if this phase is also successful, it checks the current solution against the capacity constraints on the road segments (in terms of number of vehicles that can travel through them at the same time). Whenever one sub-problem turns out to be infeasible, the algorithm backtracks and finds a different solution for the previous phase that will hopefully lead to a feasible solution for the current phase. It terminates when the last phase is feasible or all the combinations have been checked (in which case it declares the problem infeasible).    

The contributions of this paper are: (i) Designing an algorithm for the \theModel that can quickly provide feasible solutions; (ii) Compare the performance of \theAlgorithm against \MonoMod, presented in \cite{roselli2021smt} over a set of generate instances of the \theModel.

In the following,  Section~\ref{problem_formulation} provides a formal description of the problem, and Section~\ref{sec:compo_algo} introduces the algorithm together with a mathematical model and detailed descriptions of its constraints.
Finally, conclusions are drawn in Section~\ref{conclusions}.

\begin{figure*}[ht]
    \centering
    \includegraphics[width=0.9\textwidth]{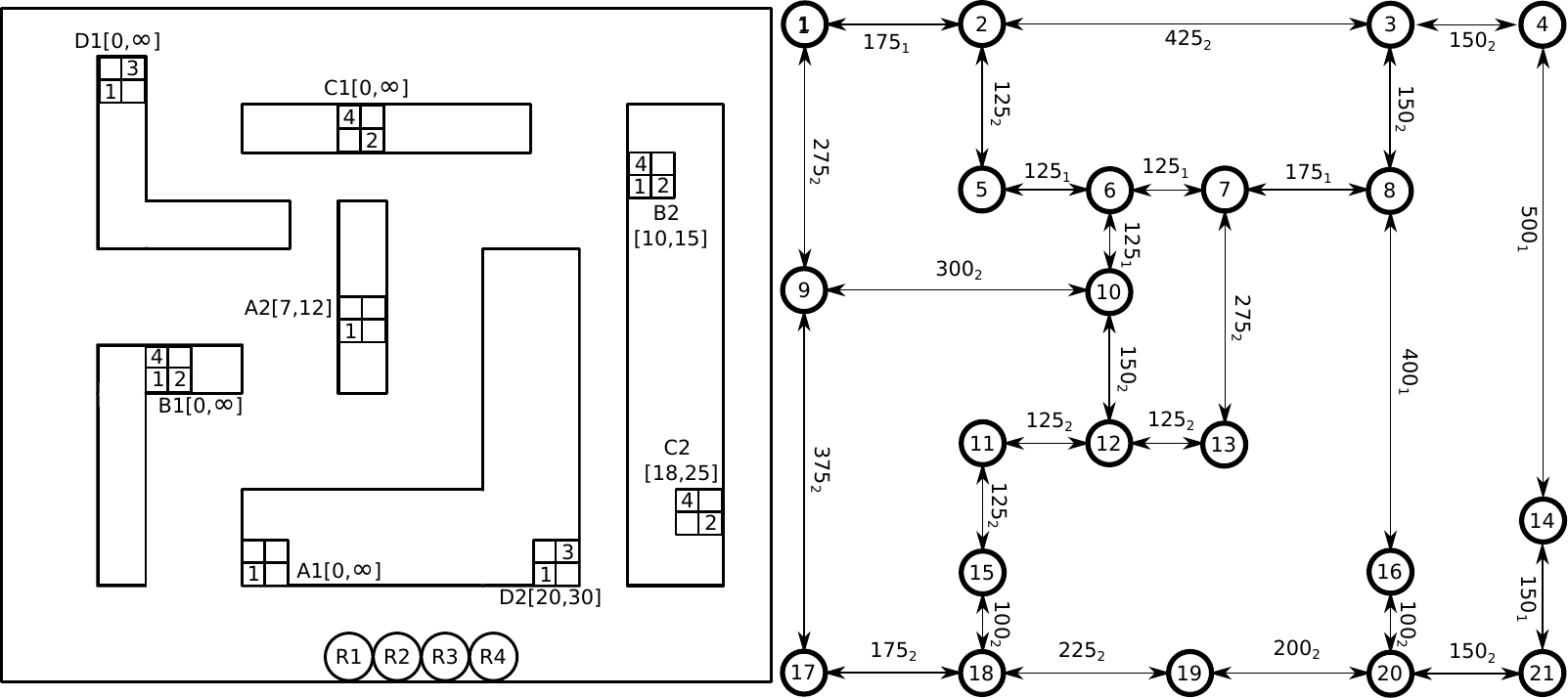}
    \caption{Problem instance of the \theModel picturing a hypothetical plant (to the left) where four AGVs (R1, R2, R3, R4), located at the bottom are available to execute four jobs (A, B, C, D), each composed by two tasks (1, 2). The plant road segments are abstracted into a strongly connected, directed, weighed graph (to the right).}
    \label{fig:example}
\end{figure*}

\section{Problem Formulation} \label{problem_formulation}

In the \theModel the plant layout is represented by a strongly connected, weighted, directed graph, where road segments are represented by the edges and the intersections between them are represented by the nodes. Customers are located at nodes and they are defined by a name (typically a number) and time window, i.e., a lower and an upper bound that represents the earliest and latest arrival allowed to serve the customer. Edges have two attributes representing respectively the road segment's length, and its capacity in terms of number of vehicles that can simultaneously travel through it. 

The following definitions are provided:
\begin{itemize}
    
    \item Task: either a pickup or a delivery operation. A task is always associated with a node (see below) where the task is executed. Each and every task has a time window as an attribute, indicating the earliest and latest time at which a vehicle can execute the task. Unless explicitly stated, the time window for a task is the time horizon.
    \begin{itemize}[label={}]
        \item $\taskSet_j, \ \forall j \in \jobSet $ : the set of tasks of job $j$
        \item $ L_{jk}, \in \nodeSet \ \forall j \in \jobSet ,k \in \taskSet_j $ : the location of task $k$ of job $j$
        \item $ \precTask_{jk} \subset \taskSet_j, \ \forall j \in \jobSet ,k \in \taskSet_j $: the set of tasks to execute before task $k$ of job $j$
        \item $ l_{jk}, \ \forall j \in \jobSet ,k \in \taskSet_j $: the time window's lower bound for task $k$ of job $j$
        \item $ u_{jk}, \ \forall j \in \jobSet,k \in \taskSet_j $: the time window's upper bound for task $k$ of job $j$
    \end{itemize}
    
    \item Job: one or more pickup tasks and one delivery task. Pickup may have precedence constraints among them, while the delivery task for a job always happens after all pickups for that job are completed. 
    
   \begin{itemize}[label={}]
        \item  \jobSet: the set of jobs
   \end{itemize}
    
    \item Vehicle: a transporter, e.g. an AGV, that is able to move between locations in the plant and perform pickup and delivery operations. 
    \begin{itemize}[label={}]
        \item  \vehicleSet: the set of all vehicles
        \item  $ \vehicleSet_j \subseteq \vehicleSet, \ \forall j \in \jobSet $: set of vehicles eligible for job $j$
        \item \OR: the maximum operating range of the vehicles
        \item \C: the charging coefficient
        \item \D: the discharging coefficient
    \end{itemize}
    
    \item Node: a location in the plant. A node can only accommodate one vehicle at the time unless otherwise specified.
    \begin{itemize}[label={}]
        \item \nodeSet: the set of nodes
    \end{itemize}

    \item Depot: a node at which one or more vehicles start and must return to after completing the assigned jobs. The depot can by definition accommodate an arbitrary number of vehicles at the same time. 
    \begin{itemize}[label={}]
        \item \Origin: the origin node
    \end{itemize}

    \item Edge: a road segment that connects two nodes.
    \begin{itemize}[label={}]
        \item $\edgeSet \subseteq \nodeSet \times \nodeSet$: the set of edges
        \item $ d_{nn'}, \ \forall n,n' \in \nodeSet$: the length of the edge connecting nodes $n$ and $n'$
        \item $ g_{nn'}, \forall n,n' \in \nodeSet$: the capacity of the edge connecting nodes $n$ and $n'$
    \end{itemize}
    
\end{itemize}
The requirements of the problem are summarized as follows:
\begin{itemize}
    \item all jobs have to be completed; for a job to be completed a vehicle has to be assigned to it and visit the locations of the job's tasks within their respective time windows.
    \item when a vehicle is assigned to more than one job, it has to execute the job's tasks sequentially; all tasks of a job must be completed before it can execute any task of another job.
    \item vehicles are powered by batteries with limited capacity but with the ability to recharge at a charging station. It is assumed that charging and discharging of the batteries is linear with respect to distance.
    \item there is only one depot, which is also the charging station; vehicles have to return to the depot they were dispatched from.
    \item different road segments in the plant have different capacities in terms of number of vehicles they can accommodate. 
    \item pickup and delivery duration (the time when the pickup and delivery tasks respectively are executed) are considered to be zero, as these times can be considered negligible compared to the travelling time.
    \item not all vehicles are eligible to execute all jobs. 
    \item It is assumed that one unit of distance is covered in one time-step, hence distance and duration are interchangeable when talking about vehicles' movements.
    \item For the algorithm to work, two additional jobs, both having  only one task located at the depot, are added: \emph{start} and \emph{end}; they are needed in the routing problem to make sure that routes begin and end at the depot.
\end{itemize}

\subsection*{Example of the \theModel}

Fig.~\ref{fig:example} shows an example of the \theModel, where four AGVs (the circles at the bottom of the plant) are available to execute four jobs, each composed by two tasks (the squares distributed over the plant). Each task is marked by an alphanumeric code where the letter refers to the job and the digit indicates the order to execute them. Next to the code, in between square brackets is indicated the time window for the execution of the tasks. The numbers inside each square indicate which AGV is eligible to execute that task. On the right, it is shown how the plant layout is abstracted into a strongly connected, directed, weighted graph (see more about this below). The nodes represent the intersections of road segments in the plant; if a task's location is close enough to an intersection, then the task will be assigned that location, otherwise a new node is added to the graph (e.g., \emph{Node 14} for \emph{task C2}). Also, \emph{Node 19} is added to the graph to locate the depot. The edges weight represent the segments' length in centimeters (regular font), and their capacity (subscript). 

The problem, described using the notation declared above, is as follows:
\begin{itemize}[label={}]
    \item $ \nodeSet = \{ 1,\ldots,21 \}$
    \item $\Origin = 19$
    \item $ \edgeSet = \{(1,2),(2,3),(3,4),(5,6),(6,7),(7,8),(9,10),$ 
    \item $ (1,12),(12,13),(17,18),(18,19),(19,20),(20,21), $ 
    \item $ (1,9),(2,5),(3,8),(6,10),(7,13),(8,16),(9,17), $ 
    \item $ (11,15),(15,18),(16,20),(14,21)  \} $
    \item $\jobSet = \{A,B,C,D \}$
    \item $\taskSet_i = \{1,2 \} \ \forall i \in \{ A,B,C,D \}$
    \item $L_{A1} = 18, \ L_{A2} = 10, \ L_{B1} = 11, \ L_{B2} = 8, $
    \item $L_{C1}=6, \ L_{C2} = 14, \ L_{D1} = 2, \ L_{D2} = 16 $
    \item $ \precTask_{A1} = \emptyset, \precTask_{A2} = {1}, \precTask_{B1} = \emptyset, \precTask_{B2} = {1}$
    \item $ \precTask_{C1} = \emptyset, \precTask_{C2} = {1}, \precTask_{D1} = \emptyset, \precTask_{D2} = {1}$
    \item $l_{A1} = 0, \ l_{A2} = 7, \ l_{B1} = 0, \ l_{B2} = 10, $
    \item $l_{C1}=0, \ l_{C2} = 18, \ l_{D1} = 0, \ l_{D2} = 20 $
    \item $u_{A1} = \infty, \ u_{A2} = 12, \ u_{B1} = \infty, \ u_{B2} = 15, $
    \item $u_{C1}=\infty, \ u_{C2} = 25, \ u_{D1} = \infty, \ u_{D2} = 30 $
    \item $\vehicleSet = \{ R1,R2,R3,R4 \}$
    \item $\vehicleSet_A = \{ R1 \}, \ \vehicleSet_B = \{ R1,R2,R4 \}, $
    \item $ \vehicleSet_C = \{ R2,R4 \}, \ \vehicleSet_D = \{ R1,R3 \} $
    \item $ \OR = 50m $
\end{itemize}

The vehicles can travel at a speed of 1m/s. A feasible schedule for the problem can be obtained by assigning \emph{R1} to \emph{A}, \emph{R4} to \emph{C}, \emph{R2} to \emph{B}, and \emph{R3} to \emph{D}.

\section{The Compositional Algorithm} \label{sec:compo_algo}

In this section we are going to introduce the sub-problems that form \theAlgorithm and are iteratively solved to find a feasible solution to the \theModel. In the implementation, each sub-problem is formulated using the SMT standard language and solved using Z3. Therefore, for each sub-problem, a set of variables is declared and as well as a set of constraints over those variables; the solver takes variables and constraints and returns an assignment for each variable such that no constraint is broken. 

The following logical operators are used to express cardinality constraints \cite{sinz2005towards} in the sub-problems:
\begin{itemize}
    \item[] $\textrm{EO}(a):$ exactly one variable of the set $a$ is true;
    \item[] $\textrm{EN}(a,n):$ exactly $n$ variables of the set $a$ are true;
    \item[] $\textrm{If}(c,o_1,o_2):$ if $c$ is \emph{true} returns $o_1$, else returns $o_2$.
\end{itemize}

The algorithm begins with a pre-processing phase where a number of paths to connect any pair of customers is computed. In theory we would like to compute \emph{all} possible paths for each pair of customers, but in practice this is not efficient because there can be too many. Also, many of these paths are not going to be used, since we are interested in the shortest ones to connect customers.

The first sub-problem aims at selecting one path for each pair of customers among all the paths computed in the pre-processing phase. This is an optimization problem whose goal is to find one path for each pair of customers such that the cumulative length of all paths is minimal. If we had computed too many paths in the pre-processing phase, this phase would be rather slow, since there can easily be millions of possible paths to connect any two customers, even for a rather small graph, and each of them is a variable for this optimization problem.  

Once we have only one path between any pair of customers, we can treat the problem as a VRPTW, with some additional constraints on the order in which we serve customers. Therefore we can set up an optimization problem to find the smallest number of routes to serve the customers within their time windows. If we cannot find a feasible set of routes, we need to go back one step and change the paths we are currently using to connect the customers. 

In the next phase, we check whether we can match the routes we computed with the available vehicles. A route may involve more than one job, and since not all vehicles are eligible to execute one route, we have to find a vehicle that can execute all jobs involved in a route. Also, routes have a latest starting time that depends on the time windows of their jobs, so we have to make sure that by that time, the vehicle selected has enough charge to finish the route. If we cannot find a feasible match, we re-run the routing problem and find another set of routes that combines the jobs differently. 

Finally, if a match between vehicles and routes can be found, the last sub-problem ensures that there is no conflict among the vehicles while they execute the routes, i.e., there is never a larger number of vehicles on a location of the graph (node or edge) than the number of vehicles allowed (based on the location's capacity).

\subsection{The Path Finder} \label{deco:path_finder}

In this phase the goal is to find a sequence of edges connecting any tasks' locations such the overall sum of distances of each path is minimized. Let $Q$ be the set of all pairs of task's locations. For any pair of points $q_i \in Q$, all possible non-cyclic paths are computed and then stored in a list $f_i$. The paths are then stored in the list \emph{Paths}: $ p_i \mapsto f_i$. 
In order to find the sequence of paths, it is possible to set up an optimization problem that uses the following decision variables:
\begin{itemize}[label={}]
    \item $\Path_{qr}$: Booleans that are \emph{True} if the $r$-th path of the $q$-th pair of points is selected.
\end{itemize}

The model is as follows:
\begin{flalign} 
    \textrm{EO}_{r\in f_q}(\textrm{\emph{path}}_{qr})  \qquad &\forall q \in Q \label{eq:exactly_one_path} \\
    \sum{\textrm{If}(\Path_{qr},|r|,0)} \qquad &\forall q \in Q, r \in f_q  \label{eq:path_finder_cost_function} 
\end{flalign}
Constraint~\eqref{eq:exactly_one_path} ensures that only one path per  pair can be selected; \eqref{eq:path_finder_cost_function} is the cost function that minimizes the overall length of the paths (in terms of nodes to visit), where $|r|$ is the length of a path $r$. 

Since the solution found may not be feasible for the following steps of the algorithm, it is necessary to store it so that it can be ruled out in the next iteration. Let $ S_{\Path} =\bigcup_{\substack{q \in Q \\ r \in f_q} } { \{ \Path^*_{qr} \} } $ be the optimal solution to the problem; also, let \emph{UsedPaths} be a list containing all the previous solutions. In order to find another feasible solution, the following constraint must be added to the model:
\begin{equation} \label{eq:previous_paths}
    \bigvee_{path_{qr} \in S_{path}}{\neg{\textrm{\emph{path}}_{qr}}} \qquad \forall S_{path} \in \textrm{\emph{UsedPaths}}
\end{equation}

Based on the model described above, it is possible to define the function \emph{pathfinder} that takes the list \emph{Paths} of all non-cyclic paths between any two points and the list \emph{UP} that contains the previously used paths as inputs, and returns \emph{CP}, the shortest feasible combination of non-cyclic paths that has not been selected yet (\emph{CP} is empty if the problem is unfeasible). 

\subsection{The Routing Problem} \label{deco:routing}

The goal is now to find feasible routes using the non-cyclic paths currently provided by the \emph{pathfinder} function to calculate the distance $d_{j_1k_1j_2k_2}$ between tasks' locations. Also, let $M_j$ be the set of mutually exclusive jobs for job $j$ (i.e. the same vehicles cannot execute job $j$ and any of the jobs in $M_j$ due to requirements on the vehicle type); let $\Perm_j$ be the set of possible orderings of tasks belonging to job $j$, where each possible ordering $\textrm{\emph{ord}}_m \in \Perm_j$ contains all tasks of job \emph{j} sorted differently.
The set of variables used to build the model for the routing problem are:
\begin{itemize}[label={}]
    \item $dir_{j_1k_1j_2k_2}$: Boolean variable that is true if a vehicle travels from the location of task $k_1$ of job $j_1$ to the location of task $k_2$ of job $j_2$
    \item $cs_{jk}$: integer variable that tracks the arrival time of a vehicle at the location of task $k$ of job $j$
    \item $rc_{jk}$: integer variable that tracks the remaining charge of a vehicle when at the location of task $k$ of job $j$
\end{itemize}
The model is as follows:
\begin{flalign}
    (cs_{jk} \geq 0 \wedge rc_{jk} \geq 0 \wedge rc_{jk} \leq \OR) \quad & \forall j\in \jobSet,k \in \taskSet_j  \label{eq:domain}\\
    \neg{dir_{jkjk}} \qquad \qquad  \forall j & \in \jobSet, k \in \taskSet_j \label{eq:not_travel_same_spot} \\
    \neg{dir_{j,k,start,k_{start}}} \qquad \qquad & \forall j \in \jobSet, k \in \taskSet_{j} \label{eq:no_travel_to_start} \\
    \neg{dir_{end,k_{end},j,k}} \qquad \qquad & \forall j \in \jobSet, k \in \taskSet_{j} \label{eq:no_travel_from_end} \\
    dir_{j_1k_1j_2k_2} \implies cs_{j_2k_2} \geq cs_{j_1k_1} + & d_{j_1k_1j_2k_2}  \nonumber \\
    \forall j_1,j_2 \in J, k_1 \in & \taskSet_{j_1}, k_2 \in \taskSet_{j_2} \label{eq:infer_arrival_time} \\
    (cs_{jk} \geq l_{jk} \wedge cs_{jk} \leq u_{jk}) \qquad \qquad & \forall j \in \jobSet, k \in \taskSet_{j} \label{eq:routing_time_window} \\
    dir_{j_1k_1j_2k_2} \implies rc_{j_2k_2} \leq rc_{j_1k_1} - & D \times d_{j_1k_1j_2k_2} \nonumber \\
    \forall j_1,j_2 \in \jobSet, k_1 \in & \taskSet_{j_1}, k_2 \in \taskSet_{j_2} \label{eq:autonomy} \\
    \textrm{EO}_{\substack{j_2 \in \jobSet \\ k_2 \in \taskSet_{j_2}}}{(dir_{j_1k_1j_2k_2})} \ \forall j_1 \in \jobSet, j_1 \neq & j_2, k_1 \in \taskSet_{j_1} \label{eq:one_arrival}\\
    \textrm{EN}_{\substack{j_2 \in \jobSet \\ k_2 \in \taskSet_{j_2}}}{(dir_{j_1k_1j_2k_2},n)} \implies \ \qquad &  \nonumber \\
    \textrm{EN}_{\substack{j_2 \in \jobSet \\ k_2 \in \taskSet_{j_2}}}&{(dir_{j_2k_2j_1k_1},n)} \nonumber \\
    \forall j_1 \in \jobSet, k_1 \in \taskSet_{j_1}, n & = 1,\dots,|\jobSet| \label{eq:flow} \\
    \bigvee_{ord_i \in \Perm_j}{\left( \bigwedge_{\substack{k_1,k_2 \in ord_i \\ k_1 \geq k_2}}{dir_{jk_1jk_2}}\right)} & \qquad  \forall j \in \jobSet \label{eq:order_within_job} \\
    \bigwedge_{P_{jk}}{cs_{jk} \geq cs_{jk'}} \qquad \qquad & \qquad \forall j \in \jobSet \label{eq:routing_precedence} 
\end{flalign}

\eqref{eq:domain} restricts the variables to be positive integers and the remaining charge to be lower than the maximum operating range; \eqref{eq:not_travel_same_spot} forbids to travel from and to the same location; \eqref{eq:no_travel_to_start} and \eqref{eq:no_travel_from_end} state that a vehicle can never travel to the start, nor travel from the end: \emph{start} and \emph{end} are physically located at the same node, but they play different roles in the routing problem, hence two different jobs; \eqref{eq:infer_arrival_time} regulates the difference in the arrival time based on the distance for a direct travel between two points; \eqref{eq:routing_time_window} enforces the time windows on the routes; \eqref{eq:autonomy} defines the decrease of charge based on the distance travelled; 
\eqref{eq:one_arrival} states that each customer must be visited exactly once; \eqref{eq:flow} guarantees the flow conservation between start and end; \eqref{eq:order_within_job} states that if a number of tasks belongs to one job, they have to take place in sequence; \eqref{eq:routing_precedence} guarantees that deliveries take place after pickups.

Finally, the cost function for the model to minimize \eqref{eq:routing_cost_function} is the total number of vehicles: 
\begin{equation} \label{eq:routing_cost_function}
    \sum{\textrm{If}(dir_{\Start,k_{\Start},j,k},1,0)} \qquad \quad \forall j \in \jobSet, k \in \taskSet_{j}
\end{equation}

If the solution of the routing problem turns out to be inconsistent with the vehicles' assignment or the conflict-free constraints in the next two phases, a new solution must be computed in order to find alternative routes for the same combination of non-cyclic paths. 
Therefore it is necessary to keep track of the combinations of routes that have already been generated so that we can rule them out when solving the routing problem again. Let $ Routes = \bigcup_{ \substack{ j \in \jobSet \\ k \in \taskSet} }{ \{ cs^*_{jk} \} } $ be the optimal solution to the routing problem found at iteration $n$ and \emph{PR} the set containing the optimal solutions found until the $(n-1)$-th iteration. In order to find the optimal set of routes, different from the ones found before, the following constraint must be added:
\begin{equation} \label{eq:previous_routes}
    \bigvee_{dir_{j_1k_1j_2k_2} \in \Routes}{\neg{dir_{j_1k_1j_2k_2}}}  \forall \Routes \in \mathit{PR}
\end{equation}

Based on the model described above, it is possible to define the function \emph{router} that takes the current combination of non-cyclic paths \emph{CP} and the set \emph{PR}, and returns a set of routes that have not been selected yet \emph{CR} (if the problem is unsat, \emph{CR} is empty). 

\subsection{The Assignment Problem} \label{deco:assignment}

The \emph{assignment problem} allocates vehicles to the routes \emph{CR} generated in the routing problem, based on the time window and eligibility requirements. In the previous phase routes were generated based only on the time windows and on the vehicles' operating range; now the actual availability of each type of vehicle is given. Moreover, the \emph{router} may generate routes that involve mutually exclusive jobs and, while it would be possible to avoid this by adding additional constraints, it would be inconvenient to do in the \emph{routing problem}, since there is no information about the vehicles assigned to the routes. On the other hand, once a set of routes is given, it can be easily checked in the \emph{assignment problem} whether a vehicle is actually eligible for a route.

Therefore, for each route $r$, we can define a list of jobs $\jobSet^r \subseteq \jobSet$ that are executed by the vehicle assigned to $r$, and the list of eligible vehicles for $r$ $El_r = \bigcap_{j\in \jobSet^r}{\vehicleSet_j}$. Also, based on the time windows of the jobs forming the routes, it is possible to work out the latest start of a route $\late_r$; for instance, if the time window's upper bound for the delivery task of a job is at time $t$ and the distance between the task's location and the depot is $d$, then the latest start is $t-d$ (remember that we assume time and distance travelled to be interchangeable). Since a route can include more than one job, the stricter deadline will define the latest start for the route.

The \emph{assignment problem} is therefore treated as a JSP where routes are jobs (whose duration depends on their length $\length_r$) and vehicles are resources, with some additional requirements on the jobs staring time. 
The set of  variables used to build the model are:
\begin{itemize}[label={}]
    \item $\allo_{ir}$: Boolean variable that is \emph{True} if vehicle $i$ is assigned to route $r$, \emph{False} otherwise ;
    \item $\Start_r$: A Non-negative integer variable that is the start time of route $r$;
    \item $\End_r$: Non-negative integer variables that keep is the end time of route $r$.
\end{itemize}
The model formulation for the assignment problem is as follows:
\begin{flalign} 
    \End_r = \Start_r + \length_r \qquad & \forall r \in R \label{eq:end_based_on_length} \\ 
    \Start_r \leq \late_r \qquad & \forall r \in R; \label{eq:latest_start} \\
    \textrm{EO}_{i\in V}{(\allo_{ir})} \qquad & \forall r \in R \label{eq:exactly_one_assignment} \\
    \bigvee_{i\in El_r}{\allo_{ir}} \qquad & \forall r \in R \label{eq:res_alloc} \\
    (\allo_{ir} \wedge \allo_{ir'}) \implies & \nonumber \\
    ( (\Start_r \geq \End_{r'} + C \cdot \length_r &) \vee \nonumber \\
    (\Start_{r'} \geq \End_{r} & + C \cdot \length_{r'})) \nonumber \\
    \forall i \in & V, r,r' \in R, r \neq r' \label{eq:non_overlap} 
\end{flalign}
\eqref{eq:end_based_on_length} connects the \emph{start} and \emph{end} variables based on the route's length; \eqref{eq:latest_start} sets the latest start time of a route based on the stricter time window among the ones of its jobs; \eqref{eq:exactly_one_assignment} states that exactly one vehicle must be assigned to a route; \eqref{eq:res_alloc} states that one (or more) among the eligible vehicles must be assigned to a route; \eqref{eq:non_overlap} states that any two routes assigned to the same vehicle cannot overlap in time; either one ends before the other starts or the other way around. 

Based on the \eqref{eq:end_based_on_length}-\eqref{eq:non_overlap} it is possible to define the function \emph{assign} that takes the routes \emph{CR} from the routing problem as input and returns \emph{AS}, which states which vehicle will drive on which route (and, of course, execute its jobs) and when it starts (if the assignment problem is unfeasible \emph{AS} will be empty).  

\subsection{The Scheduling Problem} \label{deco:scheduling}
Finally, in this phase a feasible schedule is found for the vehicles, meaning that the routes they are assigned to are checked to verify that they are conflict-free. In order to do this, we generate a list of nodes that each route visits $AN_r \ \forall r \in \Routes$, and one list of edges $AE_r \ \forall r \in \Routes$, since so far the focus was only on the tasks' locations. Note that for the same route $r$, $AE_r$ will always be one element shorter than $AN_r$ since there is an edge in between two nodes: this way the first edge in $AE_r$ is always the edge following the first node in $AN_r$, the second edge in $AE_r$ is always the edge following the second node in $AN_r$ and so on. Also, for each node in $AN_r$ it is necessary to specify whether there exist a time window, since some of the nodes are only intersection of road segments in the real plant, while others are actual pickup or delivery points). Let $l_{rn}$ and $u_{rn}$ be the earliest and latest arrival time at node $n$ on route $r$ respectively. Finally, let $\len(e)$ be the length to travel of edge \emph{e} and let $e(1)$ and $e(2)$ be the source and sink node of the edge respectively (since it is a directed graph, the edge connecting \emph{n} and \emph{n'} is different from the one connecting \emph{n'} and \emph{n}). 

This phase is also treated as a JSP, where jobs are routes, tasks within the jobs are the different nodes and edges to visit along the route while nodes and edges themselves are the resources. Also, each route has a starting time $\Start_r$ defined by the \emph{assignment model}. 
The decision variables in the \emph{scheduling problem} are:
\begin{itemize}[label={}]
    \item $\node_{rn}$: Non-negative integer variables to keep track of when route $r$ is using node $n$;
    \item $\edge_{re}$: Non-negative integer variables to keep track of when route $r$ is using edge $e$;
\end{itemize}
The model for the \emph{scheduling problem} is defined as follows:
\begin{flalign}
    \node_{rO} \geq start_r \qquad \forall r \in & \ \Routes \label{eq:route_start} \\
    \edge_{ri} \geq \node_{ri} \qquad \forall r \in & \ \Routes, i=1,\dots,|AE_r| \label{eq:visit_precedence_1} \\
    \node_{ri+1} \geq \edge_{ri} + \len(e) & \nonumber \\
    \forall r \in & \ \Routes, i=1,\dots,|AE_r| \label{eq:visit_precedence_2} \\
    (\node_{ri} \geq l_{ri} \wedge \node_{ri} \leq & u_{ri}) \nonumber \\
    \forall r \in \ & \Routes, i \in AN_r \label{eq:visit_tw} \\
    (\node_{r_1i} \geq \edge_{r_2i} + 1 \ \vee \ & \edge_{r_2i} \geq \node_{r_1i} + 1)  \nonumber \\
    \forall r_1,r_2 \in \ \Routes,  r_1 & \neq r_2, i \in AE_{r_1} \cap AE_{r_2} \label{eq:nodes_no_swap}\\
    (\edge_{r_1i} \geq \edge_{r_2i} + 1 \ \vee \ & \edge_{r_2i} \geq \edge_{r_1i} + 1) \nonumber \\
    \forall r_1,r_2 \in \Routes, r_1 & \neq r_2, i \in AE_{r_1} \cap AE_{r_2} \label{eq:edges_direct}\\
    (\edge_{r_1i_1} \geq \edge_{r_2i_2} + & \len(e_2) \ \vee \nonumber \\
    \edge_{r_2i_2} \geq & \edge_{r_1i_1} + \len(e_1)) \nonumber \\
    \forall r_1,r_2 \in R, i_1 \in AE_{r_1}, & i_2 \in AE_{r_2}, \nonumber \\
    r_1 \neq r_2, e_1(1) & = e_2(2), e_1(2) = e_2(1) \label{eq:edges_inverse}
\end{flalign}

\noindent
\eqref{eq:route_start} constraints the beginning time of a route; \eqref{eq:visit_precedence_1} and \eqref{eq:visit_precedence_2} define the precedence among nodes and edges to visit in a route; \eqref{eq:visit_tw} enforces time windows on the nodes that correspond to pickup or delivery points; \eqref{eq:nodes_no_swap} prevents vehicles for using the same node at the same time (the \emph{+1} in the constraints forbids \emph{swapping} of positions between a node and the previous or following edge) \eqref{eq:edges_direct} and \eqref{eq:edges_inverse} constraints the transit of vehicles over the same edge. If two vehicles are crossing the same edge from the same node, one has to start at least one time step later than the other and if two vehicles are traversing the same edge from opposite nodes, one has to be done transiting, before the next one can start. 

Based on the \eqref{eq:route_start}-\eqref{eq:edges_inverse} it is possible to define the function \emph{scheduler} that takes the \emph{AS} from the assignment problem as input and returns \emph{SC}, a list that expresses where each vehicle is at each time step (and, as for the previous phases, is empty if the problem is unfeasible).

\subsection{The Algorithm}

\begin{figure}
    \centering
    \includegraphics[width=0.35\textwidth]{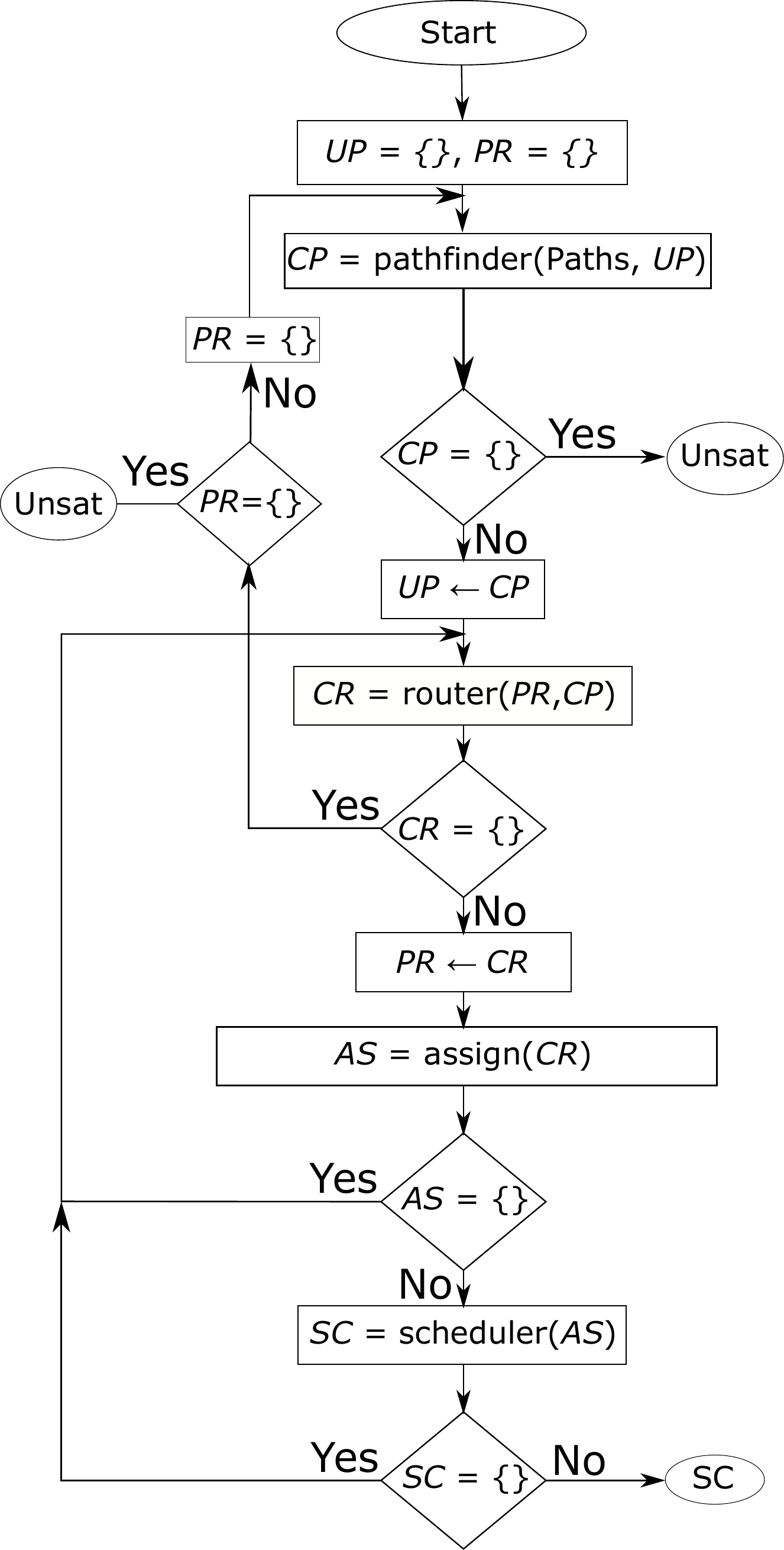}
    \caption{Flowchart of \theAlgorithm.}
    \label{fig:flowchart}
\end{figure}

In this section, the compositional algorithm to solve the \theModel is presented.
The input for the algorithm is a \emph{problem instance}, consisting of a weighted, directed \emph{graph} representing the plant layout and a list of \emph{jobs}. The function \emph{path\_enumerator} takes the graph and the jobs as input and, using Dijkstra's algorithm \cite{dijkstra1959note} returns the previously mentioned list \emph{Paths}.

The output of the algorithm is twofold: the variable \emph{solution}, which is initialized as \emph{unknown} and will possibly become either \emph{sat} or \emph{unsat}, and the \emph{schedule}, which contains the information about the location of each vehicle at each time step if the problem is \emph{sat} or is empty otherwise.

First, the algorithm finds a feasible combination of non-cyclic paths to connect any two tasks's locations: this is done through the function \emph{pathfinder}; if no combination of paths can be found (either because there are none or because all feasible solutions have been used already), the algorithm terminates and the \emph{solution} is \emph{unsat}. If a path list can be found, then such solution is added to the \emph{UP} and it is used as an input to generate feasible routes, if such exist; if no solution exists to the routing problem, there are two possible outcomes depending on the list \emph{PR}:
\begin{itemize}
    \item \emph{PR} is empty: the algorithm terminates and returns \emph{unsat};
    \item \emph{PR} is non-empty: a new combination of non-cyclic paths is computed.
\end{itemize}

\begin{table*}[htbp]
  \centering
  \caption{Comparison of \theAlgorithm and \MonoMod for the \theModel ~over a set of generated problem instances. Instances are sorted by the parameters N-V-J (nodes, vehicles, jobs), value of edge reduction, and time horizon. For each resulting class, five instances are evaluated and the number of feasible and unfeasible ones is reported, together with the average solving time (in seconds) for that specific class. The average generation time (in seconds) is also reported. The symbol ``-'' means that no instance for that category was either feasible or feasible, depending on where the symbol appears.}
  
    \resizebox{2.04\columnwidth}{!}{%
  
    \begin{tabular}{c|c|ccccc|ccccc|ccccc}
\toprule
    \multirow{3}[4]{*}{T} & \multicolumn{1}{c}{N-V-J} & \multicolumn{15}{c}{15-3-5 } \\
\cmidrule{2-17}       & \multicolumn{1}{c}{Edge Red.} & \multicolumn{5}{c|}{0} & \multicolumn{5}{c|}{25} & \multicolumn{5}{c}{50} \\
\cmidrule{2-17}       &    & Feas & Av.(sec) & Unfeas & Av.(sec) & Gen.(sec) & Feas & Av.(sec) & Unfeas & Av.(sec) & Gen.(sec) & Feas & Av.(sec) & Unfeas & Av.(sec) & Gen.(sec) \\
    \multirow{2}[0]{*}{20} & \cellcolor[rgb]{ .929,  .929,  .929}ComSat & \cellcolor[rgb]{ .929,  .929,  .929}4/5 & \cellcolor[rgb]{ .929,  .929,  .929}4.96 & \cellcolor[rgb]{ .929,  .929,  .929}1/5 & \cellcolor[rgb]{ .929,  .929,  .929}9.2 & \cellcolor[rgb]{ .929,  .929,  .929}0.36 & \cellcolor[rgb]{ .929,  .929,  .929}3/5 & \cellcolor[rgb]{ .929,  .929,  .929}6.58 & \cellcolor[rgb]{ .929,  .929,  .929}2/5 & \cellcolor[rgb]{ .929,  .929,  .929}8.9 & \cellcolor[rgb]{ .929,  .929,  .929}0.33 & \cellcolor[rgb]{ .929,  .929,  .929}1/5 & \cellcolor[rgb]{ .929,  .929,  .929}2.35 & \cellcolor[rgb]{ .929,  .929,  .929}3/5 & \cellcolor[rgb]{ .929,  .929,  .929}3.46 & \cellcolor[rgb]{ .929,  .929,  .929}0.08 \\
       & MonoMod & 4/5 & 11.05 & 1/5 & 1.16 & 22.85 & 3/5 & 8.63 & 2/5 & 4.59 & 22.94 & 1/5 & 10.98 & 4/5 & 4.24 & 22.38 \\
    \multirow{2}[0]{*}{25} & \cellcolor[rgb]{ .929,  .929,  .929}ComSat & \cellcolor[rgb]{ .929,  .929,  .929}4/5 & \cellcolor[rgb]{ .929,  .929,  .929}5.83 & \cellcolor[rgb]{ .929,  .929,  .929}1/5 & \cellcolor[rgb]{ .929,  .929,  .929}17.55 & \cellcolor[rgb]{ .929,  .929,  .929}0.33 & \cellcolor[rgb]{ .929,  .929,  .929}3/5 & \cellcolor[rgb]{ .929,  .929,  .929}6.91 & \cellcolor[rgb]{ .929,  .929,  .929}2/5 & \cellcolor[rgb]{ .929,  .929,  .929}14.14 & \cellcolor[rgb]{ .929,  .929,  .929}0.35 & \cellcolor[rgb]{ .929,  .929,  .929}3/5 & \cellcolor[rgb]{ .929,  .929,  .929}3.99 & \cellcolor[rgb]{ .929,  .929,  .929}2/5 & \cellcolor[rgb]{ .929,  .929,  .929}3.71 & \cellcolor[rgb]{ .929,  .929,  .929}0.12 \\
       & MonoMod & 4/5 & 20.79 & 1/5 & 2.18 & 41.98 & 3/5 & 24.02 & 2/5 & 33.95 & 40.88 & 3/5 & 39.42 & 2/5 & 21.7 & 43.38 \\
    \multirow{2}[0]{*}{30} & \cellcolor[rgb]{ .929,  .929,  .929}ComSat & \cellcolor[rgb]{ .929,  .929,  .929}4/5 & \cellcolor[rgb]{ .929,  .929,  .929}6.06 & \cellcolor[rgb]{ .929,  .929,  .929}1/5 & \cellcolor[rgb]{ .929,  .929,  .929}21.69 & \cellcolor[rgb]{ .929,  .929,  .929}0.36 & \cellcolor[rgb]{ .929,  .929,  .929}3/5 & \cellcolor[rgb]{ .929,  .929,  .929}7.25 & \cellcolor[rgb]{ .929,  .929,  .929}2/5 & \cellcolor[rgb]{ .929,  .929,  .929}16.49 & \cellcolor[rgb]{ .929,  .929,  .929}0.34 & \cellcolor[rgb]{ .929,  .929,  .929}3/5 & \cellcolor[rgb]{ .929,  .929,  .929}3.41 & \cellcolor[rgb]{ .929,  .929,  .929}2/5 & \cellcolor[rgb]{ .929,  .929,  .929}4.25 & \cellcolor[rgb]{ .929,  .929,  .929}0.1 \\
       & MonoMod & 5/5 & 913.15 & 0/5 & -  & 64.74 & 3/5 & 66.87 & 1/5 & 845.16 & 70.68 & 3/5 & 65.85 & 1/5 & 5.06 & 65.5 \\
    \multirow{2}[1]{*}{40} & \cellcolor[rgb]{ .929,  .929,  .929}ComSat & \cellcolor[rgb]{ .929,  .929,  .929}4/5 & \cellcolor[rgb]{ .929,  .929,  .929}6.23 & \cellcolor[rgb]{ .929,  .929,  .929}1/5 & \cellcolor[rgb]{ .929,  .929,  .929}22.97 & \cellcolor[rgb]{ .929,  .929,  .929}0.32 & \cellcolor[rgb]{ .929,  .929,  .929}3/5 & \cellcolor[rgb]{ .929,  .929,  .929}8.35 & \cellcolor[rgb]{ .929,  .929,  .929}2/5 & \cellcolor[rgb]{ .929,  .929,  .929}19.96 & \cellcolor[rgb]{ .929,  .929,  .929}0.36 & \cellcolor[rgb]{ .929,  .929,  .929}3/5 & \cellcolor[rgb]{ .929,  .929,  .929}4.05 & \cellcolor[rgb]{ .929,  .929,  .929}2/5 & \cellcolor[rgb]{ .929,  .929,  .929}4.02 & \cellcolor[rgb]{ .929,  .929,  .929}0.09 \\
       & MonoMod & 5/5 & 162.33 & 0/5 & -  & 94.36 & 3/5 & 178.07 & 0/5 & - & 98.15 & 3/5 & 240.75 & 1/5 & 901.82 & 102.1 \\
    \midrule
    \multirow{3}[4]{*}{T} & \multicolumn{1}{c}{N-V-J} & \multicolumn{15}{c}{25-4-7} \\
\cmidrule{2-17}       & \multicolumn{1}{c}{Edge Red.} & \multicolumn{5}{c|}{0} & \multicolumn{5}{c|}{25} & \multicolumn{5}{c}{50} \\
\cmidrule{2-17}       &    & Feas & Av.(sec) & Unfeas & Av.(sec) & Gen.(sec) & Feas & Av.(sec) & Unfeas & Av.(sec) & Gen.(sec) & Feas & Av.(sec) & Unfeas & Av.(sec) & Gen.(sec) \\
    \multirow{2}[0]{*}{20} & \cellcolor[rgb]{ .929,  .929,  .929}ComSat & \cellcolor[rgb]{ .929,  .929,  .929}2/5 & \cellcolor[rgb]{ .929,  .929,  .929}582.1 & \cellcolor[rgb]{ .929,  .929,  .929}3/5 & \cellcolor[rgb]{ .929,  .929,  .929}35.13 & \cellcolor[rgb]{ .929,  .929,  .929}1.24 & \cellcolor[rgb]{ .929,  .929,  .929}0/5 & \cellcolor[rgb]{ .929,  .929,  .929}- & \cellcolor[rgb]{ .929,  .929,  .929}5/5 & \cellcolor[rgb]{ .929,  .929,  .929}38.6 & \cellcolor[rgb]{ .929,  .929,  .929}1.09 & \cellcolor[rgb]{ .929,  .929,  .929}0/5 & \cellcolor[rgb]{ .929,  .929,  .929}- & \cellcolor[rgb]{ .929,  .929,  .929}5/5 & \cellcolor[rgb]{ .929,  .929,  .929}17.92 & \cellcolor[rgb]{ .929,  .929,  .929}0.84 \\
       & MonoMod & 2/5 & 42.09 & 3/5 & 26.91 & 75.04 & 0/5 & -  & 5/5 & 23.93 & 67.66 & 0/5 & -  & 5/5 & 24.38 & 70.36 \\
    \multirow{2}[0]{*}{25} & \cellcolor[rgb]{ .929,  .929,  .929}ComSat & \cellcolor[rgb]{ .929,  .929,  .929}5/5 & \cellcolor[rgb]{ .929,  .929,  .929}44.61 & \cellcolor[rgb]{ .929,  .929,  .929}0/5 & \cellcolor[rgb]{ .929,  .929,  .929}- & \cellcolor[rgb]{ .929,  .929,  .929}1.2 & \cellcolor[rgb]{ .929,  .929,  .929}4/5 & \cellcolor[rgb]{ .929,  .929,  .929}63.11 & \cellcolor[rgb]{ .929,  .929,  .929}1/5 & \cellcolor[rgb]{ .929,  .929,  .929}75.48 & \cellcolor[rgb]{ .929,  .929,  .929}1.17 & \cellcolor[rgb]{ .929,  .929,  .929}1/5 & \cellcolor[rgb]{ .929,  .929,  .929}20.99 & \cellcolor[rgb]{ .929,  .929,  .929}4/5 & \cellcolor[rgb]{ .929,  .929,  .929}227.08 & \cellcolor[rgb]{ .929,  .929,  .929}0.92 \\
       & MonoMod & 5/5 & 182.52 & 0/5 & -  & 125.76 & 4/5 & 163.15 & 1/5 & 333.66 & 109.85 & 1/5 & 110.78 & 4/5 & 113.66 & 99.82 \\
    \multirow{2}[0]{*}{30} & \cellcolor[rgb]{ .929,  .929,  .929}ComSat & \cellcolor[rgb]{ .929,  .929,  .929}5/5 & \cellcolor[rgb]{ .929,  .929,  .929}41.39 & \cellcolor[rgb]{ .929,  .929,  .929}0/5 & \cellcolor[rgb]{ .929,  .929,  .929}- & \cellcolor[rgb]{ .929,  .929,  .929}1.23 & \cellcolor[rgb]{ .929,  .929,  .929}4/5 & \cellcolor[rgb]{ .929,  .929,  .929}177.0 & \cellcolor[rgb]{ .929,  .929,  .929}1/5 & \cellcolor[rgb]{ .929,  .929,  .929}84.86 & \cellcolor[rgb]{ .929,  .929,  .929}1.13 & \cellcolor[rgb]{ .929,  .929,  .929}1/5 & \cellcolor[rgb]{ .929,  .929,  .929}34.72 & \cellcolor[rgb]{ .929,  .929,  .929}4/5 & \cellcolor[rgb]{ .929,  .929,  .929}303.96 & \cellcolor[rgb]{ .929,  .929,  .929}0.87 \\
       & MonoMod & 5/5 & 1160.64 & 0/5 & -  & 195.9 & 4/5 & 995.07 & 0/5 & -  & 169.96 & 2/5 & 479.7 & 1/5 & 1197.42 & 191.34 \\
    \multirow{2}[1]{*}{40} & \cellcolor[rgb]{ .929,  .929,  .929}ComSat & \cellcolor[rgb]{ .929,  .929,  .929}5/5 & \cellcolor[rgb]{ .929,  .929,  .929}66.63 & \cellcolor[rgb]{ .929,  .929,  .929}0/5 & \cellcolor[rgb]{ .929,  .929,  .929}- & \cellcolor[rgb]{ .929,  .929,  .929}1.11 & \cellcolor[rgb]{ .929,  .929,  .929}4/5 & \cellcolor[rgb]{ .929,  .929,  .929}139.65 & \cellcolor[rgb]{ .929,  .929,  .929}1/5 & \cellcolor[rgb]{ .929,  .929,  .929}101.52 & \cellcolor[rgb]{ .929,  .929,  .929}1.14 & \cellcolor[rgb]{ .929,  .929,  .929}1/5 & \cellcolor[rgb]{ .929,  .929,  .929}30.29 & \cellcolor[rgb]{ .929,  .929,  .929}4/5 & \cellcolor[rgb]{ .929,  .929,  .929}351.01 & \cellcolor[rgb]{ .929,  .929,  .929}0.85 \\
       & MonoMod & 4/5 & 2312.39 & 0/5 & -  & 285.38 & 4/5 & 3147.12 & 0/5 & -  & 275.71 & 2/5 & 318.09 & 0/5 & -  & 293.11 \\
    \midrule
    \multirow{3}[4]{*}{T} & \multicolumn{1}{c}{N-V-J} & \multicolumn{15}{c}{35-6-8 } \\
\cmidrule{2-17}       & \multicolumn{1}{c}{Edge Red.} & \multicolumn{5}{c|}{0} & \multicolumn{5}{c|}{25} & \multicolumn{5}{c}{50} \\
\cmidrule{2-17}       &    & Feas & Av.(sec) & Unfeas & Av.(sec) & Gen.(sec) & Feas & Av.(sec) & Unfeas & Av.(sec) & Gen.(sec) & Feas & Av.(sec) & Unfeas & Av.(sec) & Gen.(sec) \\
    \multirow{2}[0]{*}{20} & \cellcolor[rgb]{ .929,  .929,  .929}ComSat & \cellcolor[rgb]{ .929,  .929,  .929}1/5 & \cellcolor[rgb]{ .929,  .929,  .929}16.15 & \cellcolor[rgb]{ .929,  .929,  .929}4/5 & \cellcolor[rgb]{ .929,  .929,  .929}22.26 & \cellcolor[rgb]{ .929,  .929,  .929}2.46 & \cellcolor[rgb]{ .929,  .929,  .929}0/5 & \cellcolor[rgb]{ .929,  .929,  .929}- & \cellcolor[rgb]{ .929,  .929,  .929}5/5 & \cellcolor[rgb]{ .929,  .929,  .929}26.99 & \cellcolor[rgb]{ .929,  .929,  .929}2.13 & \cellcolor[rgb]{ .929,  .929,  .929}0/5 & \cellcolor[rgb]{ .929,  .929,  .929}- & \cellcolor[rgb]{ .929,  .929,  .929}5/5 & \cellcolor[rgb]{ .929,  .929,  .929}32.31 & \cellcolor[rgb]{ .929,  .929,  .929}1.77 \\
       & MonoMod & 1/5 & 80.24 & 4/5 & 97.01 & 167.36 & 0/5 & -  & 5/5 & 134.58 & 132.39 & 0/5 & -  & 5/5 & 82.83 & 131.8 \\
    \multirow{2}[0]{*}{25} & \cellcolor[rgb]{ .929,  .929,  .929}ComSat & \cellcolor[rgb]{ .929,  .929,  .929}4/5 & \cellcolor[rgb]{ .929,  .929,  .929}177.73 & \cellcolor[rgb]{ .929,  .929,  .929}1/5 & \cellcolor[rgb]{ .929,  .929,  .929}22.28 & \cellcolor[rgb]{ .929,  .929,  .929}2.32 & \cellcolor[rgb]{ .929,  .929,  .929}4/5 & \cellcolor[rgb]{ .929,  .929,  .929}81.2 & \cellcolor[rgb]{ .929,  .929,  .929}1/5 & \cellcolor[rgb]{ .929,  .929,  .929}30.37 & \cellcolor[rgb]{ .929,  .929,  .929}2.32 & \cellcolor[rgb]{ .929,  .929,  .929}4/5 & \cellcolor[rgb]{ .929,  .929,  .929}216.74 & \cellcolor[rgb]{ .929,  .929,  .929}1/5 & \cellcolor[rgb]{ .929,  .929,  .929}33.99 & \cellcolor[rgb]{ .929,  .929,  .929}1.56 \\
       & MonoMod & 4/5 & 896.66 & 1/5 & 48.93 & 220.63 & 4/5 & 636.35 & 1/5 & 32.75 & 203.98 & 4/5 & 644.79 & 1/5 & 31.24 & 201.95 \\
    \multirow{2}[0]{*}{30} & \cellcolor[rgb]{ .929,  .929,  .929}ComSat & \cellcolor[rgb]{ .929,  .929,  .929}4/5 & \cellcolor[rgb]{ .929,  .929,  .929}113.54 & \cellcolor[rgb]{ .929,  .929,  .929}1/5 & \cellcolor[rgb]{ .929,  .929,  .929}24.32 & \cellcolor[rgb]{ .929,  .929,  .929}2.27 & \cellcolor[rgb]{ .929,  .929,  .929}4/5 & \cellcolor[rgb]{ .929,  .929,  .929}188.78 & \cellcolor[rgb]{ .929,  .929,  .929}1/5 & \cellcolor[rgb]{ .929,  .929,  .929}27.8 & \cellcolor[rgb]{ .929,  .929,  .929}2.15 & \cellcolor[rgb]{ .929,  .929,  .929}4/5 & \cellcolor[rgb]{ .929,  .929,  .929}103.26 & \cellcolor[rgb]{ .929,  .929,  .929}1/5 & \cellcolor[rgb]{ .929,  .929,  .929}34.29 & \cellcolor[rgb]{ .929,  .929,  .929}1.9 \\
       & MonoMod & 4/5 & 2268.8 & 1/5 & 138.78 & 425.56 & 4/5 & 937.94 & 1/5 & 77.62 & 376.42 & 4/5 & 890.02 & 1/5 & 125.29 & 347.38 \\
    \multirow{2}[1]{*}{40} & \cellcolor[rgb]{ .929,  .929,  .929}ComSat & \cellcolor[rgb]{ .929,  .929,  .929}4/5 & \cellcolor[rgb]{ .929,  .929,  .929}216.67 & \cellcolor[rgb]{ .929,  .929,  .929}1/5 & \cellcolor[rgb]{ .929,  .929,  .929}21.65 & \cellcolor[rgb]{ .929,  .929,  .929}2.29 & \cellcolor[rgb]{ .929,  .929,  .929}4/5 & \cellcolor[rgb]{ .929,  .929,  .929}116.29 & \cellcolor[rgb]{ .929,  .929,  .929}1/5 & \cellcolor[rgb]{ .929,  .929,  .929}29.81 & \cellcolor[rgb]{ .929,  .929,  .929}2.45 & \cellcolor[rgb]{ .929,  .929,  .929}4/5 & \cellcolor[rgb]{ .929,  .929,  .929}167.9 & \cellcolor[rgb]{ .929,  .929,  .929}1/5 & \cellcolor[rgb]{ .929,  .929,  .929}32.84 & \cellcolor[rgb]{ .929,  .929,  .929}1.6 \\
       & MonoMod & 4/5 & 2689.35 & 0/5 & -  & 586.41 & 4/5 & 1167.67 & 1/5 & 913.9 & 597.01 & 4/5 & 1261.64 & 1/5 & 210.42 & 492.4 \\
    \bottomrule
    \end{tabular}%
    
    }
    
  \label{tab:table_I}%
\end{table*}%

The condition on the emptiness of \emph{PR} can save time based on one assumption: every time a new combination of non-cyclic paths is computed, it is the shortest still available. If no routing is possible, i.e. time windows could not be met with the current paths, there is no other combination of paths that will satisfy the routing problem, since they will be longer than the current one. On the other hand, if \emph{PR} is not empty, this means that routing is possible with the current combination of non-cyclic paths and it would be premature to declare the instance \emph{unsat}. Instead, the list \emph{PR} is emptied, since it only makes sense to store the old routes as long as the combination of non-cyclic paths is the same. 
If a solution to the routing problem does exist, the current routes \emph{CR} are added to \emph{PR} and then checked against the \emph{assignment problem} and the \emph{scheduling problem}. If one of these problems turns out to be unfeasible, then the function \emph{router} will look for another solution; otherwise, when both \emph{assign} and \emph{scheduler} return feasible solutions a feasible, sub-optimal schedule for the overall problem has been found. The flow chart in Fig.~\ref{fig:flowchart} shows graphically how the different sub-problems interact with each other.  

\section{Computational Analysis} \label{experiments}

In order to compare \MonoMod and \theAlgorithm we generated a set of problem instances. The parameters we used are the number of nodes, vehicles, and jobs (grouped in an index called N-J-V), as well as the time horizon and the value called 'edge reduction', which indicates the connectivity of the graph (the higher the value, the fewer edges). For each combination of these parameters, five different problems are randomly generated.

For the analysis we used Z3 4.8.9. The time limit for \MonoMod is set to 10800 seconds (three hours); the model generation time is measured separately, since it is implementation-dependent and can be dealt with using more efficient formulations, as discussed in our previous work \cite{roselli2020compact}. As for \theAlgorithm, we only computed \emph{ten} paths for each pair of customers. We also set an upper bound of \emph{ten} to the number of iterations between the \emph{Routing Problem} and the \emph{Assignment Problem}. Also, the generation time refers to the time taken to generate the paths between each pair of customers. All the experiments were performed on an \emph{Intel Core i7 6700K, 4.0 GHZ, 32GB RAM}  running \emph{Ubuntu-18.04 LTS}.

Though Z3 allows for optimization of the objective function, the size of the problems evaluated with \MonoMod is such that no optimum is expected to be found in any reasonable time. Therefore Z3 is set to find feasible, sub-optimal solutions \footnotemark.

Table~\ref{tab:table_I} summarizes the results of the computational analysis. The generation time for \theAlgorithm is actually negligible, whereas for \MonoMod it increases with N-J-V (nodes, vehicles, and jobs), and time horizon and it decreases with the \emph{edge reduction}, presumably because fewer edges means fewer constraints to declare. By comparing the number of solved instances in each category, whether they turned out to be \emph{sat} or \emph{unsat}, it is possible to notice that there is a number of instances that were determined \emph{unknown} by \theAlgorithm but were declared \emph{unsat} by \MonoMod. On the other hand, both methods usually agree on the feasibility of the \emph{sat} instances, except for some cases with high values of N-J-V and time horizon where \MonoMod run out of time. The reason for the \emph{unknown} responses lies in the termination criterion we set up for the algorithm. By running early experiments we noticed that the algorithm was rather slow in evaluating unsatisfiable problem instances; therefore we decided to limit the number of iterations between the \emph{Routing Problem} and the \emph{Assignment Problem}. On average, \theAlgorithm is between 10 and 100 times faster than the  at solving instances that are intrinsically \emph{sat}.


\section{Conclusion} \label{conclusions}

In this paper we have presented a compositional algorithm to solve the Conflict-Free Electric Vehicle Routing Problem (\theModel). We have evaluated the performance of the algorithm in handling problem instances of the \theModel and we have compared it with a monolithic model we presented in our previous work. The implementation of the model presented in Section~\ref{problem_formulation}, as well as the problem instances
are available at \hyperref[https://github.com/sabinoroselli/VRP.git]{\url{https://github.com/sabinoroselli/VRP.git}}. The algorithm proved to be significantly faster than the monolithic model to solve problem instances that are inherently satisfiable, while its performance was rather slow for unsatisfiable instances. For this reason we set up a termination criterion based on the number of iterations so that, if the algorithm cannot prove the problem either satisfiable or unsatisfiable within a certain number of iterations, it declares it \ensuremath{unknown}. We are currently working on figuring out the right conditions for the algorithm to spot unsatisfiability quicker. Also, as of today, we generate a limited number of paths to connect each pair of customers for the sub-problem \emph{Path Finder} to select one path for each pair of customers and proceed to the next sub-problems. However, in order to correctly declare a problem instance unsatisfiable, the algorithm should check all possible combinations of paths. Since the number of possible paths between two nodes can increase exponentially with the graph size, finding all paths between all pairs of nodes would be untractable. Even if finding all paths could be done instantly, having too many paths to choose from would make the \emph{Path Finder} the bottle-neck of the algorithm. Therefore we are working on an algorithm to provide the next shortest combination of paths given the current state, without enumerating all of them. 
\balance
\printbibliography{}

@article{azadeh2017robotized,
  title={Robotized Warehouse Systems: Developments and Research Opportunities},
  author={Azadeh, K and deKoster, MBM and Roy, D},
  journal={ERIM Report Series Research in Management},
  number={ERS-2017-009-LIS},
  year={2017}
}

@inproceedings{riazi2019scheduling,
  title={Scheduling and routing of {AGVs} for large-scale flexible manufacturing systems},
  author={Riazi, Sarmad and Diding, Thomas and Falkman, Petter and Bengtsson, Kristofer and Lennartson, Bengt},
  booktitle={2019 IEEE 15th International Conference on Automation Science and Engineering (CASE)},
  pages={891--896},
  year={2019},
  organization={IEEE}
}

@article{braekers2016vehicle,
  title={The vehicle routing problem: State of the art classification and review},
  author={Braekers, Kris and Ramaekers, Katrien and Van Nieuwenhuyse, Inneke},
  journal={Computers \& Industrial Engineering},
  volume={99},
  pages={300--313},
  year={2016},
  publisher={Elsevier}
}

@article{schneider2014electric,
  title={The electric vehicle-routing problem with time windows and recharging stations},
  author={Schneider, Michael and Stenger, Andreas and Goeke, Dominik},
  journal={Transportation Science},
  volume={48},
  number={4},
  pages={500--520},
  year={2014},
  publisher={INFORMS}
}

@article{krishnamurthy1993developing,
  title={Developing conflict-free routes for automated guided vehicles},
  author={Krishnamurthy, Nirup N and Batta, Rajan and Karwan, Mark H},
  journal={Operations Research},
  volume={41},
  number={6},
  pages={1077--1090},
  year={1993},
  publisher={INFORMS}
}

@inproceedings{roselli2018smt,
  title={{SMT} Solvers for Job-Shop Scheduling Problems: Models Comparison and Performance Evaluation},
  author={Roselli, Sabino and Bengtsson, Kristofer and {\AA}kesson, Knut},
  booktitle={Congress of CASE, the Portuguese Operational Research Society},
  pages={},
  year={2017},
  organization={}
}

@article{roselli2021smt,
  title={Solving the {E}lectric-{C}onflict {F}ree-{V}ehicle {R}outing {P}roblem {U}sing {SMT} {S}olvers},
  author={Roselli, Sabino and Fabian, Martin and {\AA}kesson, Knut},
  journal={MED 2021, The 29th Mediterranean Conference on Control and Automation},
  pages={},
  year={},
  organization={}
}

@article{barret_2009,
  title={Satisfiability modulo theories},
  author={Barrett, Clark W and Sebastiani, Roberto and Seshia, Sanjit A and Tinelli, Cesare and others},
  journal={Handbook of satisfiability},
  volume={185},
  pages={825--885},
  year={2009}
}

@article{DeMoura_2011,
 author = {De Moura, Leonardo and Bj{\o}rner, Nikolaj},
 title = {Satisfiability Modulo Theories: Introduction and Applications},
 journal = {Commun. ACM},
 issue_date = {September 2011},
 volume = {54},
 number = {9},
 month = sep,
 year = {2011},
 issn = {0001-0782},
 pages = {69--77},
 numpages = {9},
 url = {http://doi.acm.org/10.1145/1995376.1995394},
 doi = {10.1145/1995376.1995394},
 acmid = {1995394},
 publisher = {ACM},
 address = {New York, NY, USA}
}

@inproceedings{de_moura_2010,
  title={Z3: An efficient {SMT} solver},
  author={De Moura, Leonardo and Bj{\o}rner, Nikolaj},
  booktitle={International conference on Tools and Algorithms for the Construction and Analysis of Systems},
  pages={337--340},
  year={2008},
  organization={Springer}
}

@article{desrochers1992new,
  title={A new optimization algorithm for the vehicle routing problem with time windows},
  author={Desrochers, Martin and Desrosiers, Jacques and Solomon, Marius},
  journal={Operations research},
  volume={40},
  number={2},
  pages={342--354},
  year={1992},
  publisher={INFORMS}
}

@article{rossi2019interaction,
  title={On the interaction between Autonomous Mobility-on-Demand systems and the power network: Models and coordination algorithms},
  author={Rossi, Federico and Iglesias, Ramon and Alizadeh, Mahnoosh and Pavone, Marco},
  journal={IEEE Trans. on Control of Network Systems},
  volume={7},
  number={1},
  pages={384--397},
  year={2019},
  publisher={IEEE}
}

@article{thanos2019dispatch,
  title={Dispatch and conflict-free routing of capacitated vehicles with storage stack allocation},
  author={Thanos, Emmanouil and Wauters, Tony and Vanden Berghe, Greet},
  journal={Journal of the Operational Research Society},
  pages={1--14},
  year={2019},
  publisher={Taylor \& Francis}
}

@inproceedings{roselli2020compact,
  title={Compact Representation of Time-Index Job Shop Problems Using a Bit-Vector Formulation},
  author={Roselli, Sabino Francesco and Bengtsson, Kristofer and {\AA}kesson, Knut},
  booktitle={2020 IEEE 16th International Conference on Automation Science and Engineering (CASE)},
  pages={1590--1595},
  year={2020},
  organization={IEEE}
}

@article{braysy2005vehicle,
  title={Vehicle routing problem with time windows, Part {I}: Route construction and local search algorithms},
  author={Br{\"a}ysy, Olli and Gendreau, Michel},
  journal={Transportation science},
  volume={39},
  number={1},
  pages={104--118},
  year={2005},
  publisher={INFORMS}
}

@article{baker2003genetic,
  title={A genetic algorithm for the vehicle routing problem},
  author={Baker, Barrie M and Ayechew, MA},
  journal={Computers \& Operations Research},
  volume={30},
  number={5},
  pages={787--800},
  year={2003},
  publisher={Elsevier}
}

@article{gong2011optimizing,
  title={Optimizing the vehicle routing problem with time windows: a discrete particle swarm optimization approach},
  author={Gong, Yue-Jiao and Zhang, Jun and Liu, Ou and Huang, Rui-Zhang and Chung, Henry Shu-Hung and Shi, Yu-Hui},
  journal={IEEE Transactions on Systems, Man, and Cybernetics, Part C (Applications and Reviews)},
  volume={42},
  number={2},
  pages={254--267},
  year={2011},
  publisher={IEEE}
}

@inproceedings{sinz2005towards,
  title={Towards an optimal {CNF} encoding of boolean cardinality constraints},
  author={Sinz, Carsten},
  booktitle={International conference on principles and practice of constraint programming},
  pages={827--831},
  year={2005},
  organization={Springer}
}

@article{dijkstra1959note,
  title={A note on two problems in connexion with graphs},
  author={Dijkstra, Edsger W and others},
  journal={Numerische mathematik},
  volume={1},
  number={1},
  pages={269--271},
  year={1959}
}

@inproceedings{theunissen2018smart,
  title={Smart {AGV} system for manufacturing shopfloor in the context of industry 4.0},
  author={Theunissen, Jacobus and Xu, Hang and Zhong, Ray Y and Xu, Xun},
  booktitle={2018 25th International Conference on Mechatronics and Machine Vision in Practice (M2VIP)},
  pages={1--6},
  year={2018},
  organization={IEEE}
}

@article{carlier1989algorithm,
  title={An algorithm for solving the job-shop problem},
  author={Carlier, Jacques and Pinson, {\'E}ric},
  journal={Management science},
  volume={35},
  number={2},
  pages={164--176},
  year={1989},
  publisher={INFORMS}
}

@article{brahimi2016multi,
  title={Multi-item production routing problem with backordering: a {MILP} approach},
  author={Brahimi, Nadjib and Aouam, Tarik},
  journal={International Journal of Production Research},
  volume={54},
  number={4},
  pages={1076--1093},
  year={2016},
  publisher={Taylor \& Francis}
}

@inproceedings{bjorner_2015,
  title={{$\nu$Z}-an optimizing {SMT} solver},
  author={Bj{\o}rner, Nikolaj and Phan, Anh-Dung and Fleckenstein, Lars},
  booktitle={International Conference on Tools and Algorithms for the Construction and Analysis of Systems},
  pages={194--199},
  year={2015},
  organization={Springer}
}

\end{document}